\documentclass[pre,preprint]{revtex4-1}

\usepackage[left=1in,right=1in,top=1in,bottom=1in]{geometry}
\usepackage{graphicx}
\usepackage{epstopdf}
\usepackage{amsmath}
\usepackage{mathtools}
\usepackage{amssymb}
\usepackage{upgreek}
\usepackage{float}

\usepackage{subcaption}
\usepackage[table]{xcolor}
\usepackage[normalem]{ulem}
\graphicspath{{./figures/},{./figures_eps/}}
\usepackage{tikz}

\usepackage{verbatim}
\usepackage{ulem}
\usepackage{color,soul}
\usepackage{enumerate}

\usepackage{natbib}
\usepackage[bookmarksnumbered=true,breaklinks=true,colorlinks,citecolor=blue,linkcolor=blue]{hyperref}
\setcitestyle{numbers,square}

\newcommand{\MDGrevise}[1]{\textcolor{black}{#1}}

\newcommand{\KZrevise}[1]{\textcolor{black}{#1}}

\newenvironment{myitemize}
{ \begin{itemize}
    \setlength{\itemsep}{0pt}
    \setlength{\parskip}{0pt}
    \setlength{\parsep}{0pt}     }
{ \end{itemize}                  }

\begin{document}

\title{Data-driven control of spatiotemporal chaos with reduced-order neural ODE-based models and reinforcement learning}


\author{Kevin Zeng}
\author{Alec J. Linot}
\author{Michael D. Graham}
\email{Email: mdgraham@wisc.edu}
\affiliation{Department of Chemical and Biological Engineering, University of Wisconsin-Madison, Madison WI 53706, USA}
\date{\today}
\keywords{reduced-order modeling, reinforcement learning, spatiotemporal chaos}

\begin{abstract}
Deep reinforcement learning (RL) is a data-driven method capable of discovering complex control strategies for high-dimensional systems, making it promising for flow control applications. {In particular, the present work is motivated by the goal of reducing energy dissipation in turbulent flows, and the example considered is the spatiotemporally chaotic dynamics of the Kuramoto-Sivashinsky equation (KSE).} A major challenge associated with RL is that substantial training data must be generated by repeatedly interacting with the target system, making it costly when the  system is computationally or experimentally expensive. We mitigate this challenge in a data-driven manner by combining dimensionality reduction via an autoencoder with a neural ODE framework to obtain a low-dimensional dynamical model from just a limited data set. We substitute this data-driven reduced-order model (ROM) in place of the true system during RL training to efficiently estimate the optimal policy, which can then be deployed on the true system. {For the KSE actuated with localized forcing (``jets") at four locations}, we demonstrate that we are able to learn a ROM that accurately captures the actuated dynamics as well as the underlying natural dynamics just from snapshots of the KSE experiencing random actuations. Using this ROM and a control objective of minimizing dissipation and power cost, we extract a control policy from it using deep RL. We show that the ROM-based control strategy translates well to the {true} KSE and highlight that the RL agent discovers and stabilizes an underlying forced equilibrium solution of the KSE system. We show that this forced equilibrium captured in the ROM and discovered through RL is related to an existing known equilibrium solution of the natural KSE.
\end{abstract}

\maketitle

\section{Introduction} \label{Introduction}

In recent years, deep reinforcement learning (RL), a data-driven model-free control method, has achieved recognition for its ability to learn and discover complex control strategies for high-dimensional systems. Many of its milestone achievements involve defeating the best professional players in complex games such as GO \citep{Silver2016a}, DOTA II \citep{OpenAI2019}, and Starcraft II \citep{Vinyals2019}, as well as the leading GO, Chess, and Shogi engines \citep{Silver2018}. The successes of RL in these complex, nontrivial, and high-dimensional systems has made its application towards control of high-dimensional chaotic dynamical systems, such as turbulent flows, extremely promising. Deep RL has recently been demonstrated to be able to discover strategies in flow systems that exhibit nonchaotic dynamics such as reducing the drag in flow over bluff bodies \citep{Rabault2019,Fan2020,Tang2021,Paris2021,LiZhang2022} and decreasing convection amplitude in Rayleigh-Benard systems \citep{Beintema2020}. {Using the Kuramoto-Sivashinsky equation (KSE) as an example, } Bucci et al. {\citep{Bucci2019} demonstrated that RL is able to direct a system with natural chaotic dynamics to a given (i.e.~previously known) equilibrium solution of the system. In another study of the KSE in a chaotic parameter regime, the present authors \citep{Zeng2021} demonstrated that an RL policy whose aim is to minimize dissipation is able to discover and stabilize a nontrivial underlying equilibrium solution with low dissipation, even when no such information of its existence is given.} {The aim of the present work is to combine RL with data-driven reduced-order dynamical models for the purpose of control of spatiotemporal dynamics, as a step toward the goal of reducing energy dissipation in turbulent flows.} 

Despite {all of the successes of deep RL,} a major challenge is that it suffers from poor sampling complexity. \MDGrevise{For example,} recent work by Du et al. \citep{Du2020} showed that even if the optimal control policy could be perfectly represented by a linear function, the reinforcement learning agent still requires an exponential number of trajectories to find a near-optimal policy. Furthermore, the overall practical problem is further exacerbated by the temporal credit assignment problem, which is the difficulty in parsing out what actions in a long series of actions actually contributed to the end result--often bottlenecking RL algorithms attempting to train large models with millions of weights in application \citep{Ha2018}.

In practice this issue is often addressed by simply brute-force generating an extraordinary amount of interaction data from the target system for training. For example, to achieve the previously mentioned accolades, over 11,000+ years of DOTA II gameplay \citep{OpenAI2019}, 4.9 million self-play games of GO \citep{Silver2016a}, and 200 years of Starcraft II gameplay \citep{Vinyals2019} was required. In the realm of fluid control, even to learn a two parameter control scheme for reducing drag in an LES bluff-body simulation required over 3 weeks of training and simulation time \citep{Fan2020}. The default solution of simply generating more data can quickly make deep RL a prohibitively costly solution when the target flow system is computationally (e.g. via direct numerical simulations) or experimentally (e.g. via wind tunnel tests) expensive to realize.

Despite these challenges, there are growing efforts toward developing more data-efficient learning algorithms and work-around techniques. These efforts can be broadly categorized into three classes: advanced RL algorithms, true off-policy RL, and model-based RL. The first class, advanced RL algorithms, is comprised of methods that either build upon existing established algorithms with advanced architectures and formulations or by building in domain knowledge to improve training stability and data efficiency at the cost of increasing implementation and training complexity.

The second class, true off-policy RL, is comprised of methods that reformulate the typical semisupervised RL problem into a supervised learning problem in order to extract control strategies from an existing limited data set of transitions. However, these methods are still in the very early stages of development and it is still unclear how well they perform and generalize in complex systems. 

The last class, model-based RL, are methods that utilize a model of the target environment. These methods are advantageous in that they can leverage the existing vast classical modeling principles as well as the newly emerging deep learning techniques. As a result, model construction, implementation, and degree of data-drivenness vary widely from method to method. For example, Nagabandi et al. \citep{Nagabandi2018} introduced the Model-Based+Model-Free method, which utilizes a system model and an MPC controller to initialize the RL agent's weights. Towards more data-driven approaches, Wahlstrom et al. \citep{Wahlstrom2015} introduced Deep Dynamical Models, which uses an autoencoder to learn a reduced-order representation of the observation data and a feedforward neural network to learn a discrete time stepping model in said reduced-order representation to control an inverted pendulum from pixel inputs. Similarly, Watter et al. \citep{Watter2015} also utilized autoencoders and feedforward neural networks to learn a local linear dynamical representation of the pendulum problem. Feinberg et al. \citep{Feinberg2018} utilized models to generate short ``imagined" trajectories at each training step to better estimate the value functions. {This so-called ``Dyna"-style of model-based RL \cite{SuttonBarto2018} has been recently applied in control of spatiotemporally chaotic systems by Liu et al. \citep{Liu2021} who demonstrated with autoencoded recurrent network models containing hardcoded boundary conditions and soft conservation constraints that the number of real interactions needed with the target system to achieve control could be greatly reduced}. Finally, Ha and Schmidhuber \citep{Ha2018} introduced recurrent world models, which model the control system of interest with mixture density networks and recurrent neural networks. In this method, the authors use a data-driven model in place of the real environment during RL training to estimate a control policy for driving a 2D car and playing DOOM. We will refer to this style of utilizing a data-driven model as a surrogate training ground as ``model-based RL".

The learning process of the above defined model-based RL parallels the classic control design philosophy \citep{meyn2022} which is: 1) create a model for control, 2) design a control policy based on the model, and 3) simulate based on a high-fidelity model and repeat if needed. In parallel to these principles, we seek to: 1) \textit{learn} a model -- specifically a reduced-order, but still highly accurate model -- for control from data, 2) \textit{learn} a control policy based on the model with RL, and 3) validate based on a high-fidelity model.

\MDGrevise{A key aspect of this approach is the development of an efficient reduced-order model of the process, which we will consider to be completely governed by a state space representation 
\begin{equation}
	\frac{ds}{dt}=f(s,a),\label{eq:ODE}
\end{equation}
or its discrete-time version
\begin{equation}
	s_{t+\tau}=F(s_t,a_t),\label{eq:map}
\end{equation}
where $s\in\mathbb{R}^d$ is the state of the system and $a\in\mathbb{R}^{d_a}$ a time-dependent actuation. Our focus will be the continuous-time case Eq.~\ref{eq:ODE}, where we will seek a reduced-order representation
\begin{equation}
	\frac{dh}{dt}=f(h,a),\label{eq:redODE}
\end{equation}
where $h\in\mathbb{R}^{d_h}$ and $d_h\ll d$. 
 Before describing our specific approach, we briefly review other methods for developing dynamic models from data. A more detailed discussion is given in \cite{Linot2021}.}


\MDGrevise{If the governing equations of the process are known, one approach is simply to use a direct simulation without any model reduction. Here of course none of the efficiency gains enabled by reducing the dimension of the model are realized.  A traditional approach to reducing the dimension of these problems is the Galerkin projection method \cite{Holmes2012} and other variants like nonlinear Galerkin \cite{Marion1990,Jauberteau1990,Graham1993} or posprocessing Galerkin \cite{GarciaArchilla1998}, which have higher accuracy than the traditional Galerkin approach for given number of unknowns.  Additionally, such approaches can be improved by adding data-driven information. For example, Wan et al. \cite{Wan2018} approximated the unresolved dynamics in nonlinear Galerkin with an LSTM. The concept of combining first-principles and data-driven models is a widely studied one (see e.g.~``physics-informed neural networks'' \cite{Raissi2019}). 
}

\MDGrevise{In the present work, however, our focus is data-driven models. The classical approach is to seek a linear discrete-time model; a specific example of this approach, which has seen wide use recently even for nonlinear systems, is dynamic mode decomposition (DMD)\cite{DMDBook}. We describe and implement an adaption of DMD to closed loop control, known as DMDc, in Section IIIB. As a linear state space model, at long times, trajectories predicted by DMD always evolve toward a fixed point or a quasiperiodic orbit with a discrete set of frequencies, so it cannot faithfully characterize chaotic dynamics, which is our interest here. In principle, for all nonlinear systems there does exist a linear, albeit infinite-dimensional, operator called the \emph{Koopman} operator that describes the evolution of arbitrary observables \cite{Lasota.1994.10.1007/978-1-4612-4286-4,Budisic2012}. We choose not to take this approach because our aim is to reduce the dimension of our model. In any case, effective methods for approximation of chaotic long-time dynamics with data-driven Koopman operator approximations remain elusive.}

\MDGrevise{With the increasing power of neural network representations, nonlinear data-driven modeling methods have received substantial recent attention. Two popular methods include recurrent NN (RNN) \citep{Hopfield1982, Schmidhuber1997} and reservoir computing \citep{Jaeger2001, Maass2007}. These approaches are similar in that a hidden or reservoir state $r_t$ is used along with the current state $s_t$ to predict a future hidden or reservoir state $r_{t+\tau}$. Then a function is used to predict the new state $s_{t+\tau}$ from the hidden state.  The two approaches are different in how these functions are approximated. In reservoir computing the functions for evolving the reservoir are chosen a priori, and the mapping back to $s_{t+\tau}$ is learned. In RNNs, specific neural networks are trained to learn all of the functions --  two examples are LSTMs \citep{Schmidhuber1997} and gated recurrent units \cite{Chung2014}. Vlachas et al.~\cite{Vlachas2019} shows that both these methods perform well at predicting chaotic dynamics. However, because they involve hidden variables, these approaches do not reduce the dimension of the system, but rather increase it: i.e.~they do not determine $s_{t+\tau}$ from $s_t$ alone, but rather also use past history stored in $r_t$ to make predictions.  } 


\MDGrevise{A more direct approach to learning dynamical models is to
	learn the right-hand side (RHS) of Eq.~\ref{eq:ODE} or Eq.~\ref{eq:map} from data. 
One way to do this, at least for low-dimensional systems, is using the ``Sparse Identification of Nonlinear Dynamics '' (SINDy) approach \cite{Brunton2016}. 
In this method a dictionary of candidate nonlinear functions are selected to approximate $f$ in Eq.~\ref{eq:ODE}. Then, by using sparse regression, the dominant terms are identified and kept. 
Alternately, one could simply represent $f$ with a neural network \cite{Gonzalez-Garcia1998}. If one does not have time-derivative data or estimates, then the neural ODE approach \citep{Chen2018,Linot2021} can be used. } 


\MDGrevise{Now we turn to the issue of dimension reduction.} The dynamics of formally infinite-dimensional partial differential equations describing dissipative flow systems are known to collapse onto a finite-dimensional invariant manifold \citep{Teman1977}--a so-called ``inertial manifold". \MDGrevise{In past works, it was shown that one can capture a high-dimensional system's dynamics on this lower-dimensional manifold by using a combination of autoencoders, to identify the manifold coordinates, and Neural ODEs (NODE) \citep{Chen2018}, to model the dynamics on the manifold in discrete time \cite{Linot2020} or continuous time } \citep{Linot2021}.

Our aim is to adapt this data-driven, low-dimensional model framework as a data-efficient surrogate training grounds for model-based RL. Once training is complete, this ROM-based policy is then deployed to the real system for assessment or further fine tuning. {We refer to our method as \textit{Data-driven Manifold Dynamics for Reinforcement Learning}, or DManD-RL for short.}


Unlike the previous implementations of data-driven models in model-based RL \citep{Ha2018,Watter2015,Wahlstrom2015}, which model the evolution of the controlled system in fixed discrete time steps, our method models the vector field of the dynamics on the manifold in the presence of control inputs using neural ODEs \citep{Chen2018}. The choice of using neural ODEs has several benefits: 1) it allows one to utilize the vast and already present array of integration schemes for time-stepping, 2) it allows one to train the model from data collected from variable time step sizes as well as make variable time step predictions, which grants one the ability to adjust the transition time step in RL without resorting to recollecting new data and training a new dynamics model, 3) it is a natural formulation for the modeling of the dynamics of our physical flow systems. We highlight that \MDGrevise{this framework is not limited to fixed time intervals}, can utilize unevenly { and/or widely} spaced data in time, respects the Markovian nature of the \MDGrevise{dynamics of the systems we are interested in capturing}, and respects the Markov environment assumption that underlies RL theory \citep{SuttonBarto2018}, unlike other common data-driven approaches such as recurrent neural networks \citep{Hopfield1982, Schmidhuber1997} and reservoir computing \citep{Jaeger2001, Maass2007}.  \KZrevise{Finally, we do not impose that the underlying relationship between the control input and the dynamics is affine in control} (\MDGrevise{i.e. that~$\frac{ds}{dt}=f(s)+g(s)a$ where $s$ is the state and $a$ is the control input), as some past works have done \citep{Watter2015}. This is advantageous for application towards complex flow control systems of practical interest as \MDGrevise{ many, such as aircraft  \citep{Boskovic2004} and underwater vehicles \citep{Geranmehr2015}}, do not have relationships with this affine structure.}


The remainder of this paper is structured as follows: In Section \ref{ProblemSetUp} we introduce the Kuramoto-Sivashinsky equation, a proxy system for turbulent flows that displays rich spatiotemporal chaos, and the control objective which serves as a drag reduction analogue. We conclude this section with an outline of our ROM-based RL framework. In Section \ref{Results} we examine the performance of our learned reduced-order model, the control strategy extracted from the model using RL, and the dynamical systems relevancy of the strategy. In Section \ref{Conclusions} we conclude our findings and discuss applications to more complex systems.

\section{Formulation} \label{ProblemSetUp}
\subsection{Kuramoto-Sivashinsky Equation}

The Kuramoto Sivashinsky Equation (KSE) is given by,

\begin{equation}
  \frac{\partial v}{\partial t} = -v\frac{\partial v}{\partial x} -\frac{\partial^2 v}{\partial x^2} -\frac{\partial^4 v}{\partial x^4} + f(x,t).
 \label{eq:KSE}
\end{equation}

Here $f$ is a spatio-temporal forcing term that will be used for control actuation. We consider the KSE in a domain of length $L=22$ with periodic boundary conditions as this is a dynamically well characterized system \citep{Cvitanovic2010}. The uncontrolled KSE, $f=0$, exhibits rich dynamics and spatio-temporal chaos, which has made it a common toy problem and proxy system for the Navier-Stokes Equations. Spatially localized control is implemented in the KSE with $N=4$ equally spaced Gaussian ``jets" located at $X\in\{0,L/4,2L/4,3L/4\}$ as done by Bucci et al. \citep{Bucci2019} and Zeng and Graham \citep{Zeng2021}, where $a_i(t)\in[-1,1]$ is the control signal output by the RL control agent,

\begin{equation}
  f(x,t) = \sum_{i=1}^{4}\dfrac{a_i(t)}{\sqrt{2\pi\sigma_s}}\exp\left(-\dfrac{(x-X_i)^{2}}{2\sigma_s^{2}}\right).
 \label{eq:forcingterm}
\end{equation}

The system is time evolved with a time step of $\Delta t=0.05$ using the same numerical method and code as Bucci et al. \citep{Bucci2019} with a third-order semi-implicit Runge-Kutta scheme, which evolves the linear second and fourth order terms with an implicit scheme and the nonlinear convective and forcing terms with an explicit scheme. Spatial discretization is performed with Fourier collocation on a mesh of 64 {evenly spaced }points {and in our formulation the state vector $u$ consists of the solution values at the collocation points.}    To serve as an analogue to energy-saving flow control problems, we are interested in the minimization of the dissipation, $D$, and total power input, $P_f$ required to power the system and jets of the KSE system, which are described by $D = \langle (\frac{\partial^2 u}{\partial x^2})^2\rangle$ and $P_f = \langle (\frac{\partial u}{\partial x})^2\rangle+ \langle uf \rangle$, respectively. Here $\langle \cdot \rangle$ is the spatial average. 

\subsection{Background and Method Formulation}
\begin{figure}[t]
	\begin{center}
		\includegraphics[width=0.50\textwidth]{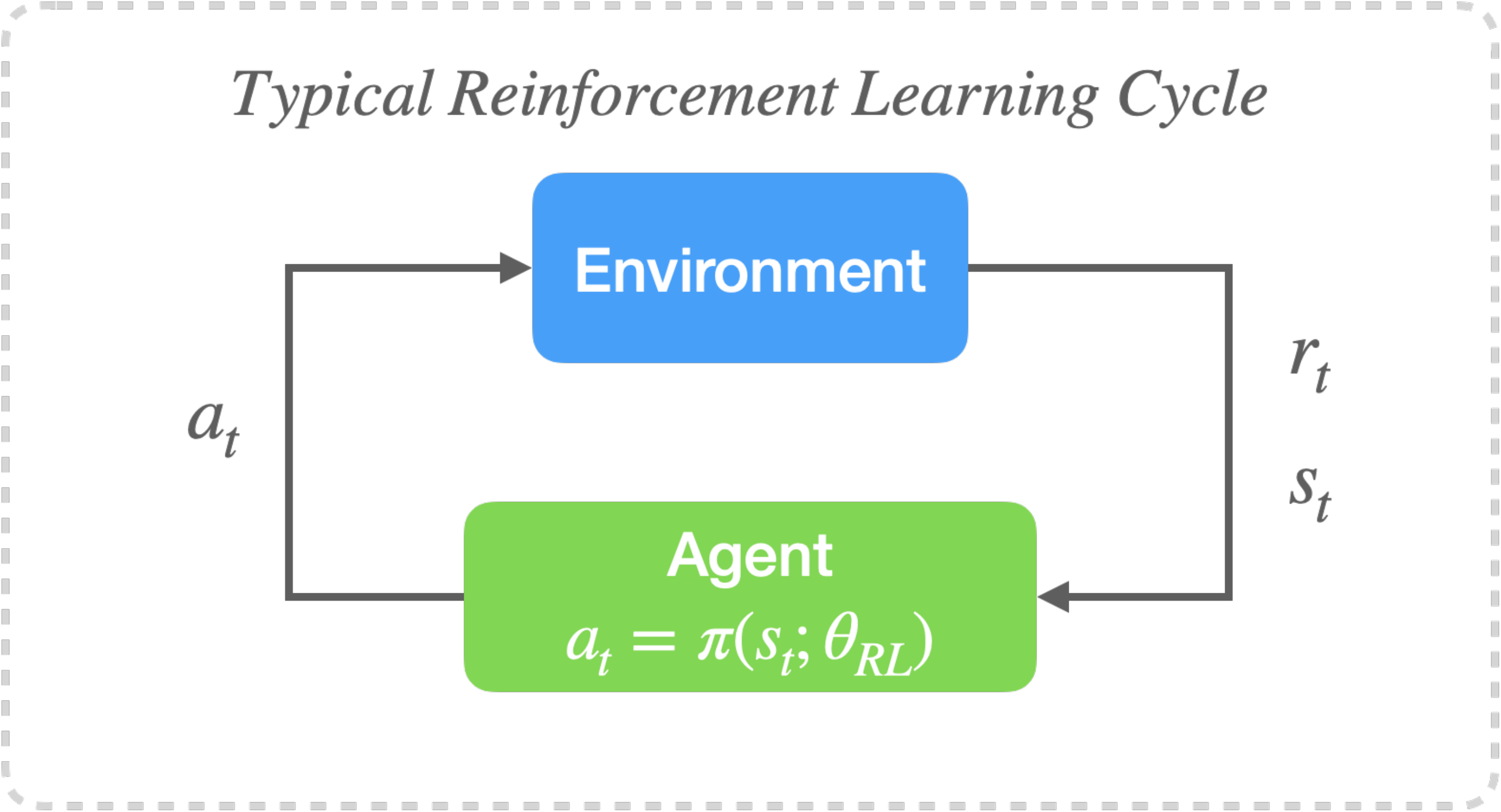}
		\caption[]{Typical reinforcement learning cycle.}
		\label{fig:RLSchematic}
	\end{center}
\end{figure}

At its core, deep RL is a cyclic learning process with two main components: the RL agent and the environment. The agent is the embodiment of the control policy and generally a neural network, while the environment is the target system for control. During each cycle, the agent makes a state observation of the system, $s_t$, and outputs an action, $a_t$. The impact of this action on the environment is then quantified by a scalar quantity, $r_t$, which is defined by the control objective. During training, the agent attempts to learn the mapping between $s_t$ and $a_t$ that maximizes the cumulative long time reward and updates accordingly each cycle. This learning cycle is shown in Fig. \ref{fig:RLSchematic}.

\begin{figure}[t]
	\begin{center}
		\includegraphics[width=0.95\textwidth]{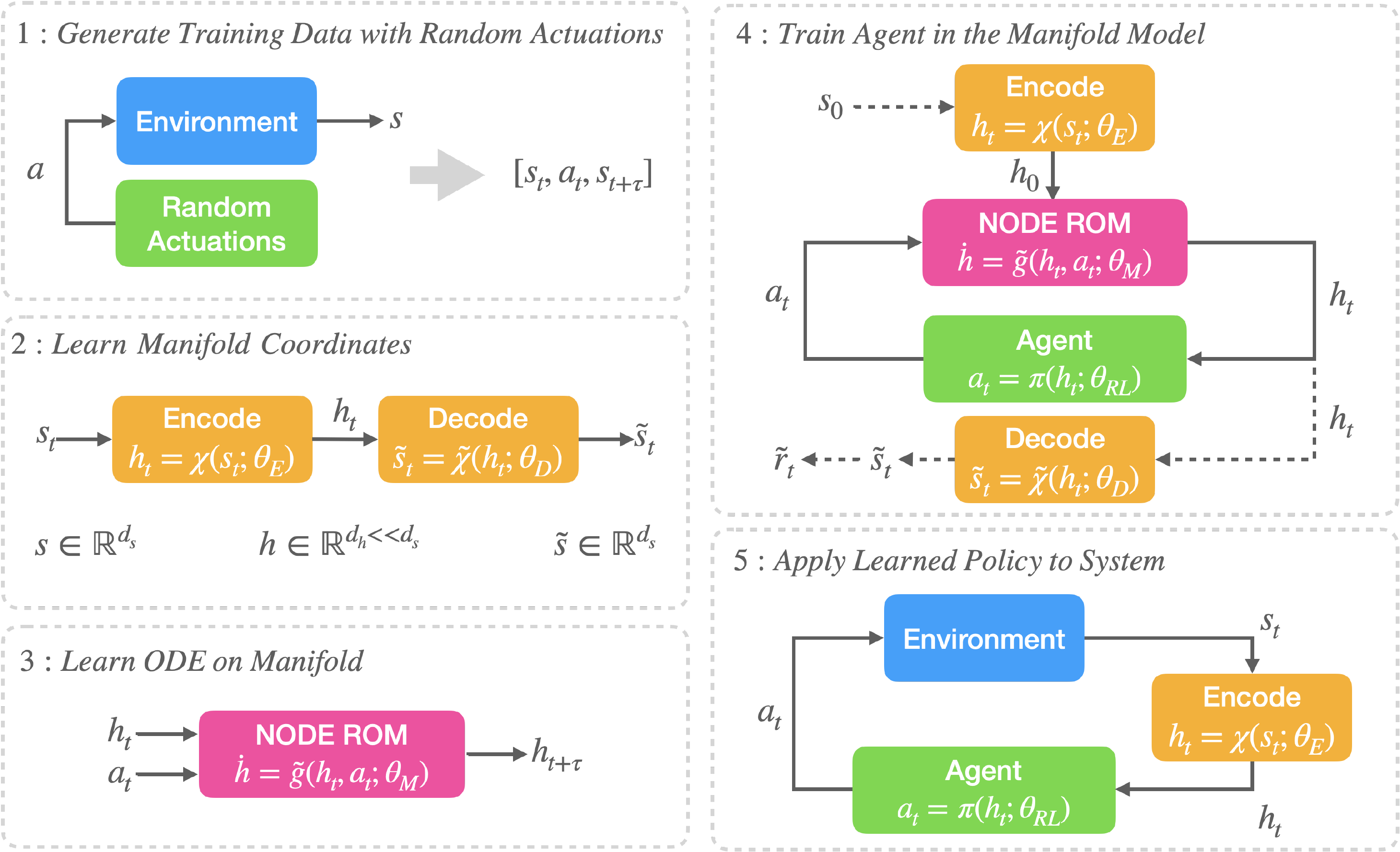}
		\caption[]{Procedure for learning a NODE-ROM from data, combining it with RL to approximate a control policy, and deploying the approximate policy back to the true system. }
		\label{fig:ROMSchematic}
	\end{center}
\end{figure}

Deep RL's poor sampling complexity makes this cyclic process costly or even intractable for learning control strategies when simulations or experiments of the environment are expensive to realize. We aim to circumvent this issue in a completely data-driven manner by training the RL agent with a surrogate reduced-order model (ROM) of the target system. Our method can be divided into five steps: 1) obtain ROM training data 2) learn a reduced-order embedding coordinate transformation using an autoencoder 3) learn the RHS of the controlled system's dynamics in the reduced-order coordinates using Neural ODEs, 4) extract a control policy from this ROM using deep RL, 5) deploy and assess the control strategy in the real system. An outline of this process is presented in Fig. \ref{fig:ROMSchematic}.

\textbf{Step 1: Training Data Collection:}
We obtain our training data for our data-driven model by observing actuated trajectories of the target system. As we assume there is neither a model nor control strategy available at this stage, the actuations applied are randomly sampled from a uniform distribution of the available range of control inputs. The observed series of states, $s_t$, applied actions, $a_t$, at time $t$, and the resulting state after $\tau$ time units, $s_{t+\tau}$ are saved as transition snapshots $[s_t,a_t,s_{t+\tau}]$ for ROM training. In our demonstration with the KSE, we take the state observable to be the \MDGrevise{solution} of the KSE at time $t$, $s_t = u(t)$, and the coinciding control signals to the jet actuations to be the action, $a_t = a(t)$ (where $a_i(t)$ is the signal to the $i$th jet). 

\textbf{Step 2: Learning the Manifold Coordinate System:}
We utilize undercomplete neural network autoencoders \citep{Goodfellow2016} to learn the coordinate transformation to and from the lower-dimensional manifold of the actuated dynamics of our system. {Undercomplete autoencoders are hourglass-shaped neural networks that are composed of two subnetworks, the encoder and decoder, are connected by a size-limiting bottleneck layer that explicitly restricts the number of degrees of freedom that the input data must be represented by. This structure forces the encoder to learn to compress the input data to a representation that fits through the bottleneck and the decoder to learn to reconstruct the original input from the compressed representation while minimizing information loss across the network. In our work,} the encoder, $h_t=\chi(s_t;\theta_{E})$, is tasked with learning the mapping from the full-state representation, $s_t\in\mathbb{R}^{d_s}$, to the manifold representation, $h_t\in\mathbb{R}^{d_h}$, and is parameterized by weights $\theta_{E}$. Importantly, $d_h \ll d_s$, such that the bottleneck layer explicitly restricts the number of degrees of freedom representing our data,  as our objective is to learn a reduced-order manifold representation. The decoder, $\hat{s_t}=\tilde{\chi}(h_t;\theta_{D})$, is tasked with learning the mapping from the manifold representation to the full-state representation and is parameterized by weights $\theta_{D}$. The autoencoder is trained to minimize the mean squared reconstruction loss (MSE) $\mathcal{L}_{AE}=\langle ||s_t-\hat{s_t}||^2 \rangle$ where $\hat{s_t}=\tilde{\chi}(\chi(s_t))$ using the snapshots of state data obtained from the previous step and $\langle\cdot\rangle$ is the average over a training batch. In our example with the KSE we also perform an intermediary change of basis from the observed state $s_t$ to its projection onto the Principal Component Analysis basis (computed from the training data) prior to input to $\chi(s_t;\theta_{E})$ and a return to the full space post output from $\tilde{\chi}(h_t;\theta_{D})$ as we found this to be an effective intermediate basis in learning. We identify the dimension of the finite dimensional manifold, $d_\mathcal{M}$, by tracking the MSE performance of the autoencoders as we vary $d_h$, \citep{Linot2020,Linot2021}.

\textbf{Step 3: Learning the Actuated Dynamics along the Manifold:}
We next develop a neural-ODE (NODE) model of {the dynamics in the manifold coordinates $h$ by learning the RHS (vector field) of the ODE $\dot{h}=g(h_t,a_t;\theta_M)$, where $\theta_M$ denotes the neural-network parameters that are to be determined \citep{Chen2018}.} To make dynamical forecasts given a manifold state, $h_t$, and input control action, $a_t$, the evolution of the manifold state $\tau$ time units forward, $h_{t+\tau}$, can be computed {for a given $g(h_t,a_t;\theta_M)$ with a standard time-integration algorithm}. To train the NODE network, the same dataset generated in step 1, $[s_t,a_t,s_{t+\tau}]$ can used. Because the NODE network's purpose is to model the vector field of the dynamics in the manifold coordinates, we must first converted the training data to $[h_t,a_t,h_{t+\tau}]$ using $\chi(s_t;\theta_{E})$. This data set can then be used to train the NODE network to minimize the forecasting loss: $\mathcal{L}_{NODE}=\langle ||h_{t+\tau}-\hat{h}_{t+\tau}||^2 \rangle$ where $\hat{h}_{t+\tau}$ is the forecast made by integrating the NODE with a numerical integrator forward $\tau$ time units. {In this work we use the Dormand-Prince 45 numerical scheme \citep{Dormand1980}. The gradient of the loss with respect to the network parameters, $\theta_M$, can be obtained by either performing back-propagation via automatic differentiation through all the steps of the time integration scheme (which can be memory intensive for very large forecasts) or by the adjoint method described by Chen et al. \citep{Chen2018}. In this work we opt for the former as we did not encounter memory issues and past works have demonstrated good agreement between both options \citep{Linot2021}.} We highlight that once training is complete, NODE forecasted trajectories, which are in the manifold coordinates, can be recovered back to the original ambient space, $s$, at any point using the decoder, $\tilde{\chi}(h_t;\theta_{D})$, obtained in step 2.


\textbf{Step 4: Learning a control strategy from the NODE-ROM with deep RL:}
We differentiate our {DManD-RL} method from typical RL with two distinctions. First the RL agent learns by interacting with the learned NODE-ROM, not the true environment. Second, during training the RL agent learns in the manifold coordinate, $h$, not the observable space, $s$. {We point out that the usual RL nomenclatures for state transitions, $s_{t+1}$ or $s'$,  are written here as $h_{t+\tau}$ in the manifold representation and $s_{t+\tau}$ in the ambient representation, to make explicit the fact that the time interval $\tau$ is in fact a parameter of the system.} To train the RL agent, the NODE-ROM is first initialized with an encoded initial condition. The agent learns by interacting only with the NODE-ROM: given a state in the manifold representation, $h_t$, the agent attempts to map it to the optimal control action. This action is then passed back to the NODE-ROM and the evolution in $h$ subject to the prescribed actions is obtained by integrating the NODE-ROM forward in time.  In this work we aim to lower the total power consumption $D+P_f$ of the KSE dynamics, so we define  the \KZrevise{immediate} reward for the RL algorithm as $r_t=-(D(t)+P_f(t))$.  The algorithm seeks to maximize the long time discounted cumulative reward $R_t=\sum_{l=0}^{\infty} \gamma^{l}r_{t+l\tau}$ where $\gamma=0.99$. The reward return for this manifold state-action pair ($[h_t,a_t]$) can be estimated by decoding the resulting manifold trajectory using $\tilde{\chi}(h_t;\theta_{D})$ and estimating $r_t$, which is used to update the agent. The learning cycle then repeats. In this work we utilize the Deep Deterministic Policy Gradient (DDPG) RL method \citep{lillicrap2019}, but we emphasize that any general RL method can be used with our framework. 


\textbf{Step 5: Deploying and Validating the \KZrevise{DManD-RL} Control Strategy:}
 Once RL training within the NODE-ROM is complete, the learned \KZrevise{DManD-RL} policy can be \MDGrevise{applied} to the true system. As the agent was trained in the reduced manifold space, the encoder obtained in step 2 must be inserted between the environment and agent to map state observations to the manifold representation prior to input to the agent. The \KZrevise{DManD-RL} policy can then be applied in typical closed-loop fashion. If desired, new additional on-policy data can be collected to further improve the model/agent in an iterative fashion or the agent can even be simply fine-tuned with real-system training.

\section{Results} \label{Results}

\subsection{Neural ODE Model Evaluation}
\begin{figure}[t]
 	\centering
	\begin{subfigure}[t]{0.9\textwidth}
		\includegraphics[width=\textwidth]{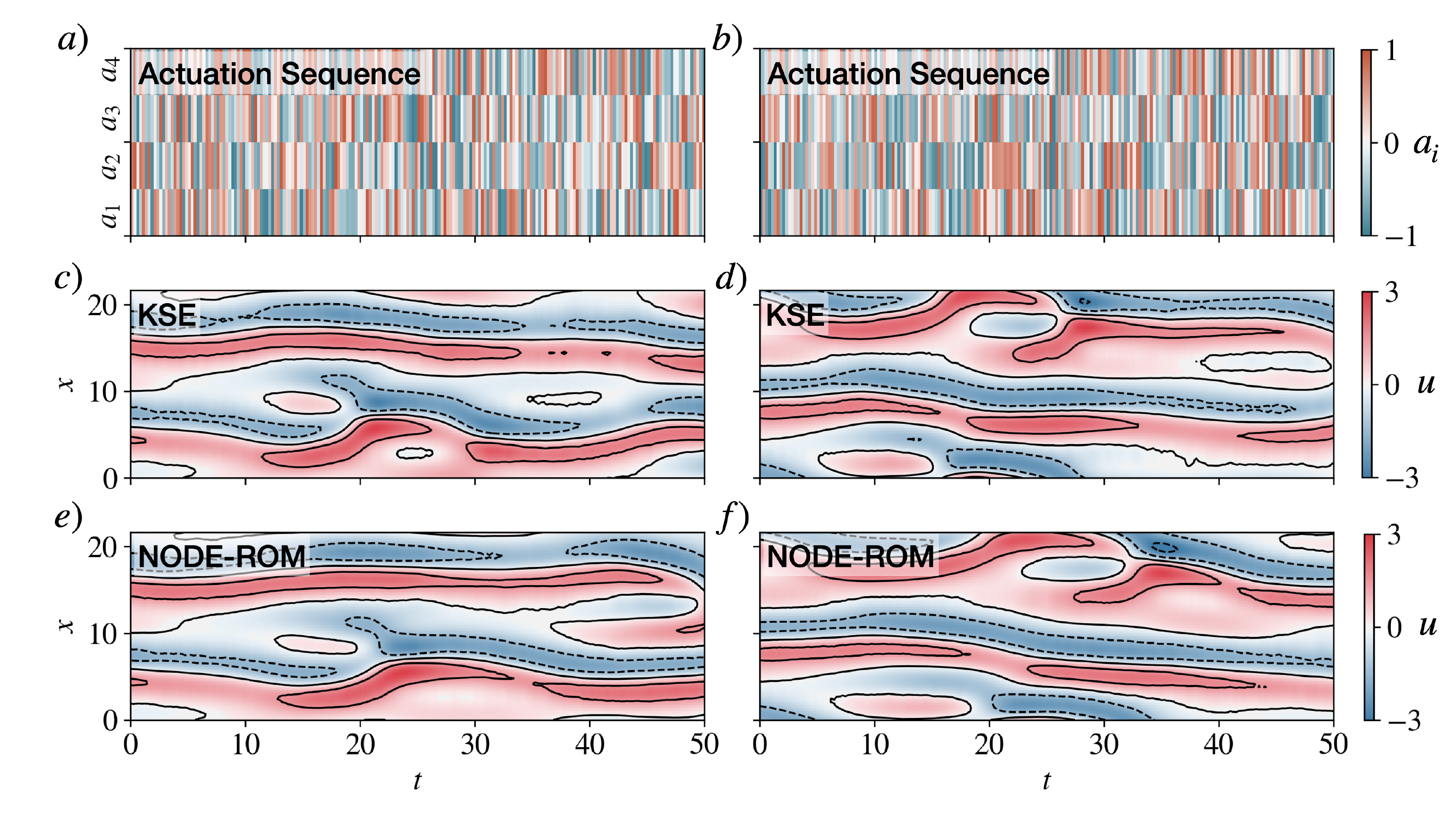}\phantomcaption\label{fig:KSE_ExampleActuatedTraj-a}
	\end{subfigure}
	\begin{subfigure}[t]{0.0\textwidth}
		\includegraphics[width=\textwidth]{KSE_Actuated_Traj}\phantomcaption\label{fig:KSE_ExampleActuatedTraj-b}
	\end{subfigure}
	\begin{subfigure}[t]{0.0\textwidth}
		\includegraphics[width=\textwidth]{KSE_Actuated_Traj}\phantomcaption\label{fig:KSE_ExampleActuatedTraj-c}
	\end{subfigure}
	\begin{subfigure}[t]{0.0\textwidth}
		\includegraphics[width=\textwidth]{KSE_Actuated_Traj}\phantomcaption\label{fig:KSE_ExampleActuatedTraj-d}
	\end{subfigure}
	\begin{subfigure}[t]{0.0\textwidth}
		\includegraphics[width=\textwidth]{KSE_Actuated_Traj}\phantomcaption\label{fig:KSE_ExampleActuatedTraj-e}
	\end{subfigure}
	\begin{subfigure}[t]{0.0\textwidth}
		\includegraphics[width=\textwidth]{KSE_Actuated_Traj}\phantomcaption\label{fig:KSE_ExampleActuatedTraj-f}
	\end{subfigure}
	\caption[]{(a) random \MDGrevise{actuation} sequences $a_i(t)$ (c) ground truth KSE trajectory starting from a random initial conditions following actuation sequences in (a), (e) the decoded NODE-ROM trajectory following actuation sequence (a) and the same initial condition in (c). A second example is shown in (b), (d), and (f), respectively. }
	\label{fig:KSE_ExampleActuatedTraj-full}
\end{figure}

Training data was collected by capturing snapshots of the KSE experiencing jet actuations whose magnitude and direction are randomly sampled from a uniform distribution of the allowable action range described in Section \ref{ProblemSetUp}. We collect 40,000 of these transition snapshots spanning 10,000 time units for our NODE-ROM training, where $\tau=0.25$ time units. We emphasize here that in this study $\tau$ is kept constant for simplicity but in practice $\tau$ can vary from snapshot to snapshot. Actuation signals applied during these trajectories and during control are held constant for the duration of $\tau$ (i.e. zero-order hold; see Fig. \ref{fig:KSE_ExampleActuatedTraj-a} for typical actuation trajectories). 

As the autoencoder serves as the mapping function to and from the manifold representation $h_t$, we need to estimate the number of degrees of freedom in the manifold, $d_h$, to accurately capture the dynamics of the system. As $d_h$ is generally not known a priori for complex systems, we empirically estimate it by tracking the reconstruction error as a function of $d_h$ as done by Linot and Graham \citep{Linot2020,Linot2021}. In this work our autoencoder networks utilize encoders and decoders each with a hidden layer of size 500 activated by sigmoid functions.

\begin{figure}[t]
 	\centering
	\begin{subfigure}[t]{0.9\textwidth}
		\includegraphics[width=\textwidth]{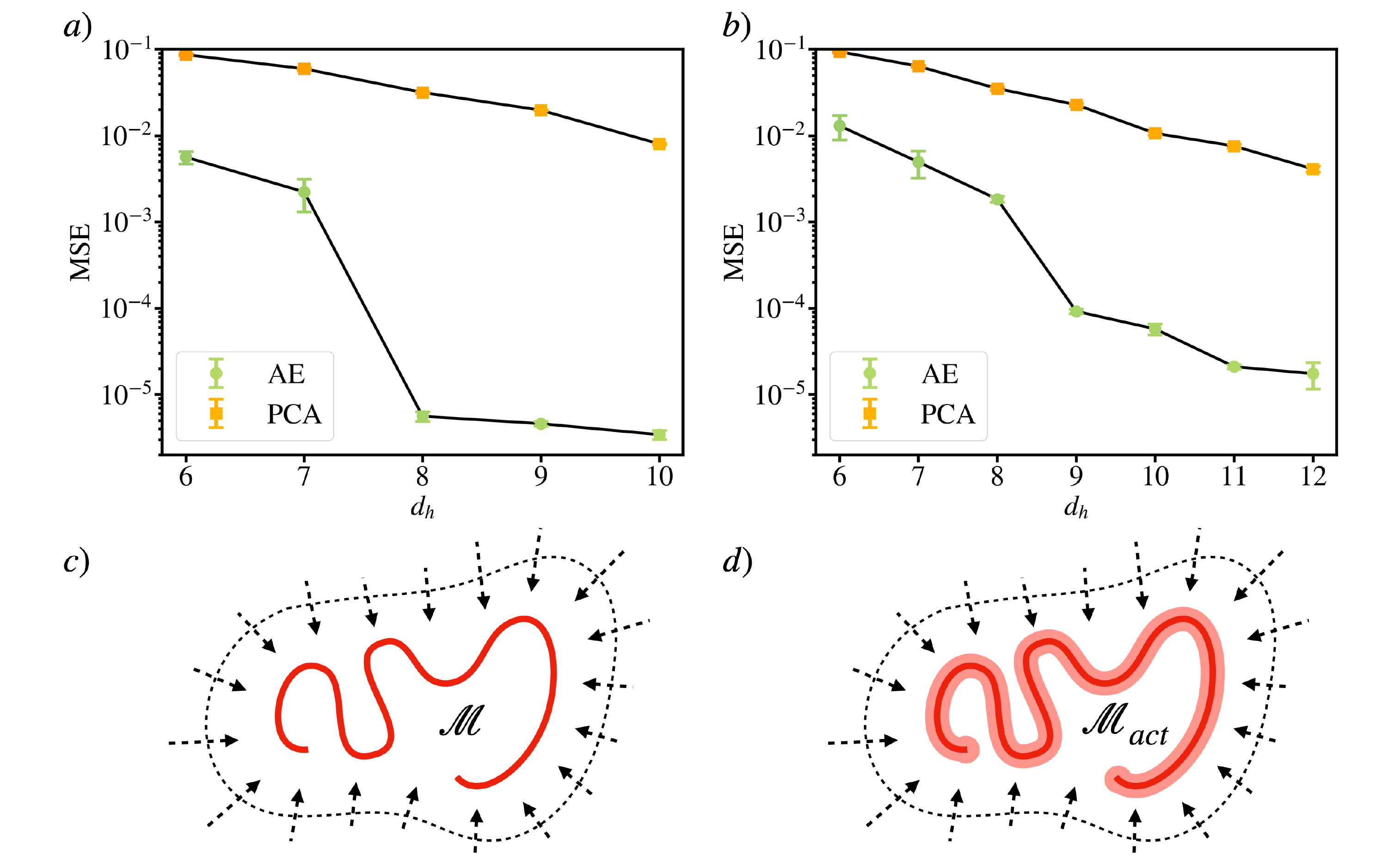}\phantomcaption\label{fig:KSE_Auto_MSE-a}
	\end{subfigure}
	\begin{subfigure}[t]{0.0\textwidth}
		\includegraphics[width=\textwidth]{KSE_Auto_MSE}\phantomcaption\label{fig:KSE_Auto_MSE-b}
	\end{subfigure}
	\begin{subfigure}[t]{0.0\textwidth}
		\includegraphics[width=\textwidth]{KSE_Auto_MSE}\phantomcaption\label{fig:KSE_Auto_MSE-c}
	\end{subfigure}
	\begin{subfigure}[t]{0.0\textwidth}
		\includegraphics[width=\textwidth]{KSE_Auto_MSE}\phantomcaption\label{fig:KSE_Auto_MSE-d}
	\end{subfigure}
	\caption[]{Mean squared reconstruction error for autoencoders trained on a) unactuated and b) actuated KSE data vs. manifold representation dimension, $d_h$. Cartoon illustrations of the inertial manifolds in which the data lives on for the c) unactuated and d) actuated system.}
	\label{fig:KSE_Auto_MSE-full}
\end{figure}

\KZrevise{Before considering actuated data, we first apply the autoencoder analysis to the natural unactuated KSE.  For the domain size $L=22$ considered here, the manifold dimension has been determined to be $d_\mathcal{M}=8$ in past studies \citep{Ding2016,Linot2020}. This will aid in validating our procedure and give insight into the results with the actuated system.}
To validate our procedure with the natural KSE, we collect 40,000 snapshots of the KSE where $a_t=0$, to capture the unactuated KSE dynamics. Shown in Fig. \ref{fig:KSE_Auto_MSE-a} are the mean-squared errors (MSE) of reconstruction for undercomplete autoencoder networks trained on natural, unactuated KSE snapshots for  bottleneck layers of varying dimension $d_h$. We note that as $d_h$ increases from 7 to 8 the MSE exhibits a sharp drop. Here, $d_h=8$ corresponds to the minimum number of degrees of freedom required to fully represent the manifold in which the natural KSE data lives,  which agrees with previous studies \citep{Linot2020, Ding2016}. We denote this manifold dimension of $d_h=8$ as $d_{\mathcal{M}}$. We note that increasing $d_h$ beyond $d_{\mathcal{M}}$ does not significantly improve reconstruction error, indicating that the data can be effectively represented in $d_h=8$ and additional dimensions are superfluous \citep{Linot2020}.

Repeating this analysis with the snapshots collected from the randomly actuated KSE to estimate the dimensionality of the actuated dynamics, we observe a similar phenomenon. Shown in Fig. \ref{fig:KSE_Auto_MSE-b} are the mean-squared errors of reconstruction as $d_h$ is changed for undercomplete autoencoders trained with KSE snapshots obtained from trajectories perturbed randomly by the four jet actuators. We note that compared to the Fig. \ref{fig:KSE_Auto_MSE-a}, the sharp drop in MSE is delayed to $d_h=9$, with a lesser drop at $d_h=11$.  

To dynamically explain this, shown in Fig. \ref{fig:KSE_Auto_MSE-c} is a cartoon of the $d_\mathcal{M}=8$ embedded manifold $\mathcal{M}$ on which the long time dynamics of the unactuated KSE lives, where states that begin off of the manifold are attracted exponentially to it due to dissipation. When actuations are applied, the KSE produces states and trajectories that live off of the natural attractor, effectively giving the manifold  ``thickness" in additional dimensions, \MDGrevise{as} shown schematically in Fig. \ref{fig:KSE_Auto_MSE-d}. This ``thickness" require additional degrees of freedom, i.e. increased $d_{\mathcal{M}}$ from the original system, to accurately capture the dynamics. 

\begin{figure}[t]
	\begin{center}
		\includegraphics[width=0.7\textwidth]{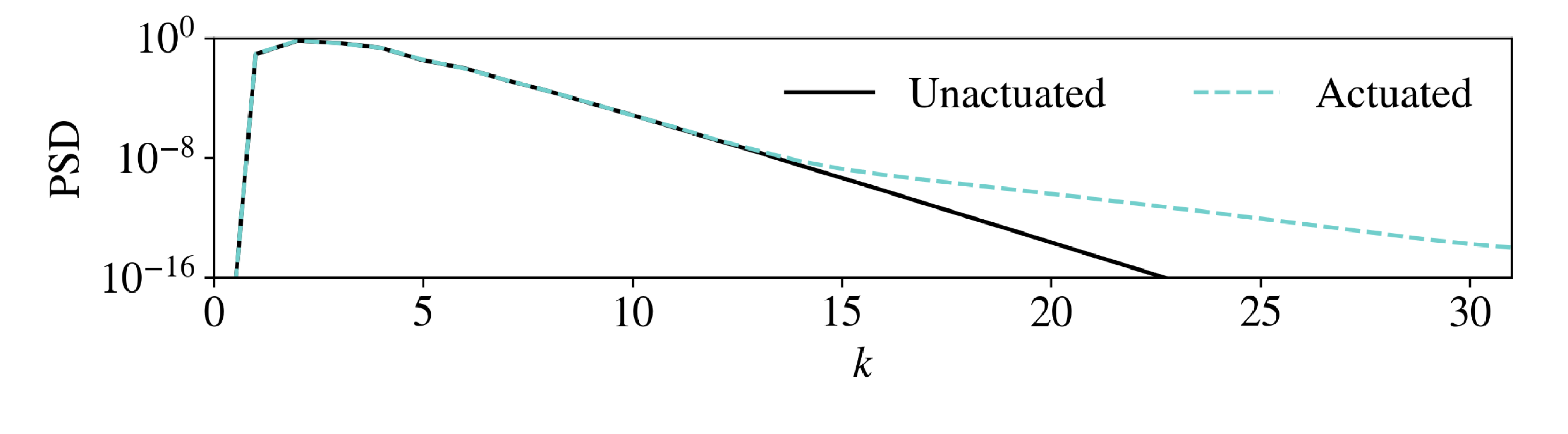}
		\caption[]{Power spectral density vs.~wavenumber of unactuated and actuated data.}
		\label{fig:KSE_PSD}
	\end{center}
\end{figure}

To further support this view, shown in Fig. \ref{fig:KSE_PSD} is the power spectral density of the unactuated and actuated data sets $u(t)$ used to train the autoencoders shown in Fig. \ref{fig:KSE_Auto_MSE-full}. We note that in the presence of actuations the data exhibits a broadening of the high-wavenumber tail compared to the unactuated data. Physically, this indicates that there are additional high-wavenumber spatial features in the actuated data set, which require additional degrees of freedom to accurately capture. Finally, we highlight in Fig. \ref{fig:KSE_Auto_MSE-a} and Fig. \ref{fig:KSE_Auto_MSE-b} that our autoencoders outperform dimensionality reduction using only Principal Component Analysis in reconstruction error by several orders of magnitude once $d_{\mathcal{M}}$ has been reached. 

The autoencoder MSE landscape of the actuated KSE data suggests that a manifold representation of $d_h=12$ should be sufficient to capture the dynamics of the controlled KSE. Using the $\chi,\tilde{\chi}$ mappings learned by the $d_h=12$ autoencoder, a NODE network with two hidden layers of size 200, 200 with sigmoid activation was trained to model the actuated dynamics in the $d_h=12$ manifold. This NODE was trained using the same data used to train the autoencoder by simply converting the collected data of $[s_t,a_t,a_{t+\tau}]$ to $[h_t,a_t,h_{t+\tau}]$ with $h_t=\chi(s_t;\theta_{E})$.

To demonstrate the learned NODE-ROM's predictive capability, shown in Fig. \ref{fig:KSE_ExampleActuatedTraj-c} and Fig. \ref{fig:KSE_ExampleActuatedTraj-d} are two example KSE trajectories experiencing random jet actuation signal sequences visualized in Fig. \ref{fig:KSE_ExampleActuatedTraj-a}, Fig. \ref{fig:KSE_ExampleActuatedTraj-b}, respectively. We also show in Fig. \ref{fig:KSE_ExampleActuatedTraj-e} and Fig. \ref{fig:KSE_ExampleActuatedTraj-f} the decoded $d_h=12$ trajectory forecast by the NODE-ROM beginning from the same initial condition and following the same actuation signal sequence as its ground truth trajectory, Fig. \ref{fig:KSE_ExampleActuatedTraj-c}, Fig. \ref{fig:KSE_ExampleActuatedTraj-d}, respectively. We note that the forecasted trajectories agree qualitatively with the ground truth for about 20-30 times units or 1-1.5 Lyapunov times of the natural system. 

\begin{figure}[t]
 	\centering
	\begin{subfigure}[t]{0.9\textwidth}
		\includegraphics[width=\textwidth]{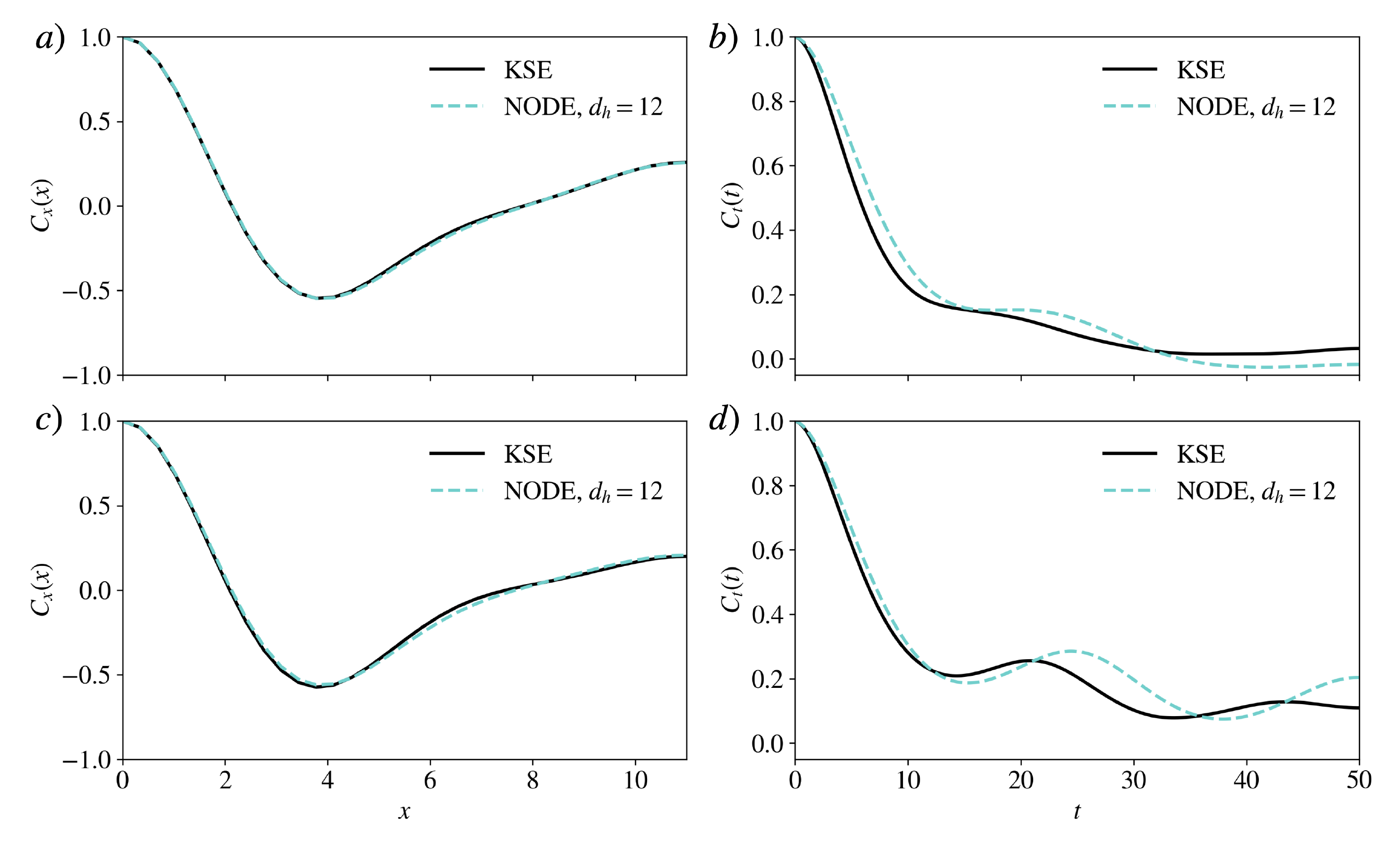}\phantomcaption\label{fig:KSE_ActuatedCorr-a}
	\end{subfigure}
	\begin{subfigure}[t]{0.0\textwidth}
		\includegraphics[width=\textwidth]{KSE_Corr_Combined}\phantomcaption\label{fig:KSE_ActuatedCorr-b}
	\end{subfigure}
	\begin{subfigure}[t]{0.0\textwidth}
		\includegraphics[width=\textwidth]{KSE_Corr_Combined}\phantomcaption\label{fig:KSE_UnactuatedCorr-c}
	\end{subfigure}
	\begin{subfigure}[t]{0.0\textwidth}
		\includegraphics[width=\textwidth]{KSE_Corr_Combined}\phantomcaption\label{fig:KSE_UnactuatedCorr-d}
	\end{subfigure}
	\caption[]{a) Spatial and b) temporal autocorrelation computed from an ensemble of trajectories experiencing random jet actuations. NODE-ROM ($d_h=12$) forecasts were produced using the same initial conditions and jet actuation sequences as each respective trajectory produced from the true KSE. c) Spatial and d) temporal autocorrelation computed from an ensemble of trajectories with actuations set to zero. NODE-ROM ($d_h=12$) forecasts were produced using the same initial conditions as each respective trajectory produced from the true KSE. }
	\label{fig:KSE_Corr-full}
\end{figure}

To quantitively compare the ensemble performance of the NODE-ROM and the actuated KSE, an ensemble of 50 actuated forecast/ground truth pairs of trajectories (where the random actuation sequence and initial conditions between each pair are the same) was used to compute the spatial and temporal autocorrelation, shown in Fig. \ref{fig:KSE_ActuatedCorr-a} and Fig. \ref{fig:KSE_ActuatedCorr-b}, respectively. We note that the NODE-ROM accurately captures the spatial autocorrelation of the actuated KSE, while the temporal autocorrelation exhibits good agreement with a slight temporal dilation. These results indicate that the NODE-ROM is accurately capturing the distribution of features of the actuated KSE in both space and time.

\begin{figure}[t]
 	\centering
	\begin{subfigure}[t]{0.9\textwidth}
		\includegraphics[width=\textwidth]{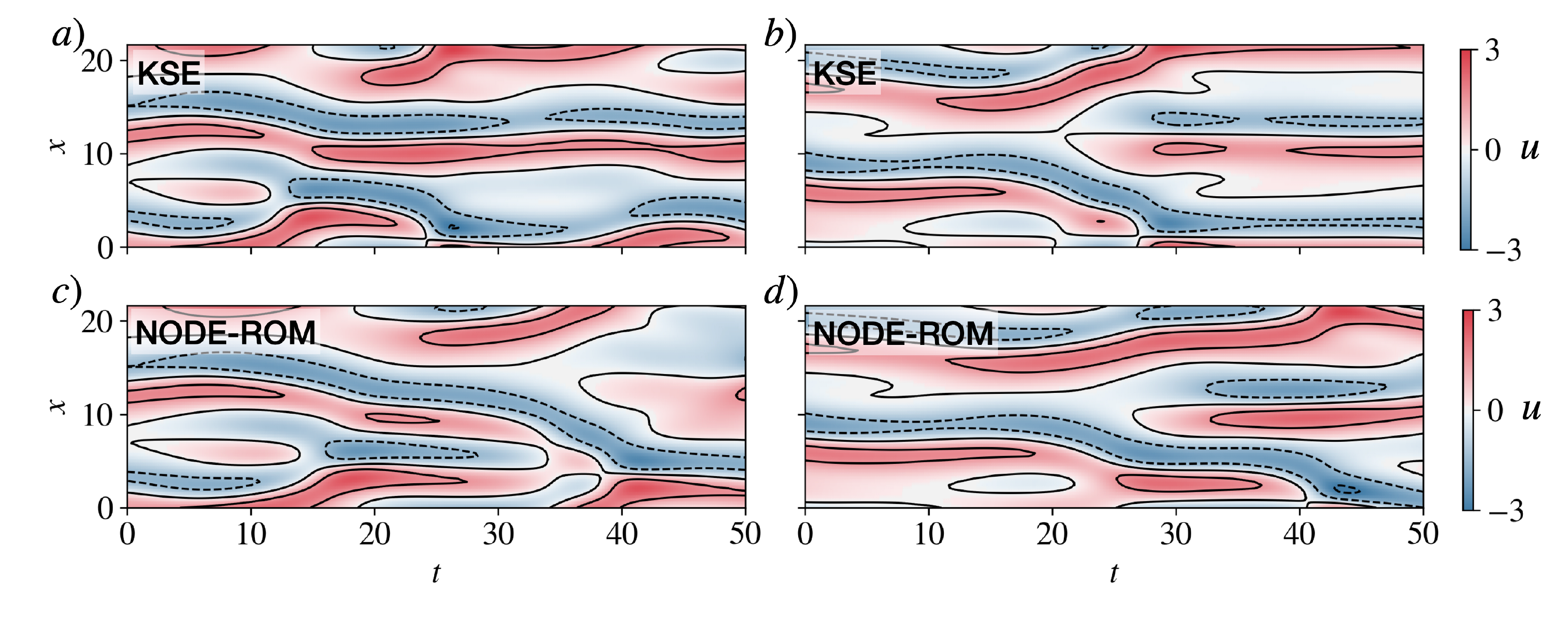}\phantomcaption\label{fig:KSE_ExampleUnactuatedTraj-a}
	\end{subfigure}
	\begin{subfigure}[t]{0.0\textwidth}
		\includegraphics[width=\textwidth]{KSE_Unactuated_Traj}\phantomcaption\label{fig:KSE_ExampleUnactuatedTraj-b}
	\end{subfigure}
	\begin{subfigure}[t]{0.0\textwidth}
		\includegraphics[width=\textwidth]{KSE_Unactuated_Traj}\phantomcaption\label{fig:KSE_ExampleUnactuatedTraj-c}
	\end{subfigure}
	\begin{subfigure}[t]{0.0\textwidth}
		\includegraphics[width=\textwidth]{KSE_Unactuated_Traj}\phantomcaption\label{fig:KSE_ExampleUnactuatedTraj-d}
	\end{subfigure}
	\caption[]{Example unactuated trajectories of the KSE in a) and b) and their corresponding decoded NODE-ROM ($d_h=12$) forecasts starting from the same initial conditions in c) and d), respectively.}
	\label{fig:KSE_ExampleUnactuatedTraj-full}
\end{figure}

As the NODE-ROM was trained with only transition snapshots experiencing random actuations, a natural question is:  how well does our model capture the underlying natural dynamics, i.e.~with $a_t=0$? Shown in Fig.\ref{fig:KSE_ExampleUnactuatedTraj-a} and Fig.\ref{fig:KSE_ExampleUnactuatedTraj-b} are two example natural KSE trajectories where $a_t=0$. Accompanying the unactuated ground truth trajectories shown in Fig.\ref{fig:KSE_ExampleUnactuatedTraj-a} and Fig.\ref{fig:KSE_ExampleUnactuatedTraj-b} are the decoded $d_h=12$ trajectories forecasted by the same NODE-ROM that was trained on actuated data beginning from the same initial conditions and with zero actuation signal input, shown in Fig.\ref{fig:KSE_ExampleUnactuatedTraj-c} and \ref{fig:KSE_ExampleUnactuatedTraj-d}, respectively. We note that the forecasted trajectories agree qualitatively with the ground truth for about 1-1.5 Lyapunov times. We again assess the spatial and temporal autocorrelations between an ensemble of 50 pairs of  ground truth trajectories and NODE-ROM forecasts, shown in Fig. \ref{fig:KSE_UnactuatedCorr-c} and \ref{fig:KSE_UnactuatedCorr-d}, respectively. We note that the ROM matches the spatial autocorrelation of the KSE very well while the temporal autocorrelation reveals that the ROM exhibits a more pronounced {but still quantitatively small} temporal dilation. We emphasize that despite the training data for the ROMs were obtained from randomly actuated trajectories and that there is no substantial unactuated data, the ROMs still recovers the unactuated dynamics well.


\subsection{Model-Based Control Performance}
With this data-driven NODE-ROM, a control agent was trained by interacting only with the NODE-ROM in the manifold space with $r_t$ estimated from the $s=\tilde{\chi}(h;\theta_{D})$ decoded state.  
 We set the NODE-ROM model transition time to be $\tau=0.25$ for all experiments.
 The control agent was trained with 1000 episodes of 100 time units  long (i.e. 400 transitions per episode), with each episode beginning from a random on-attractor initial condition of the natural, {i.e.~unforced}, KSE. Jet actuations implemented by the control agent, $a_t$, were maintained constantly from $s_t$ to $s_{t+\tau}$. In this work the DDPG actor and critic networks utilized ReLU activated hidden layers of size 128 and 64, {respectively}, followed by tanh and linear activations  to the outputs of size 4 and 1, respectively. 

\begin{figure}[t]
 	\centering
	\begin{subfigure}[t]{0.9\textwidth}
		\includegraphics[width=\textwidth]{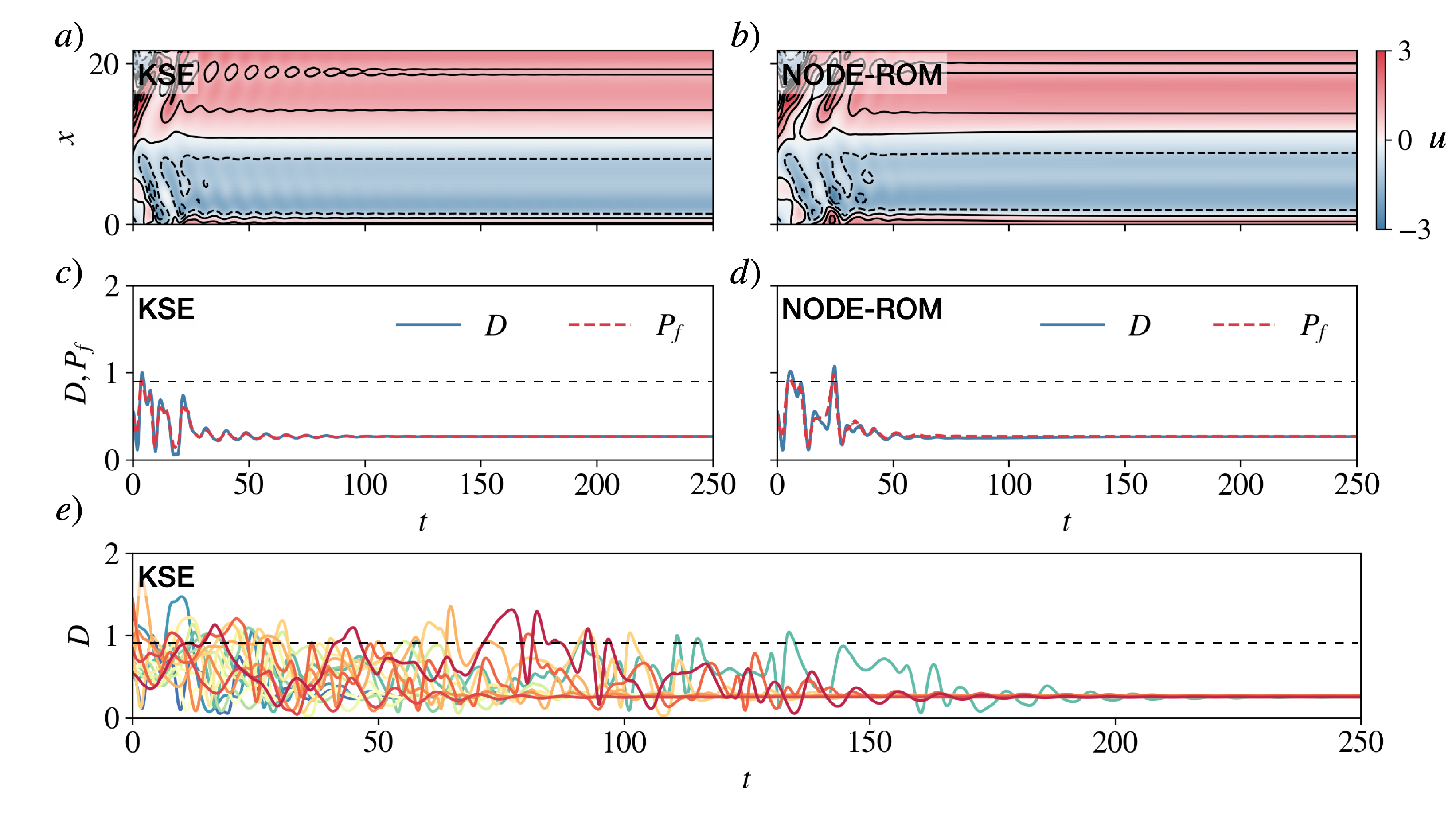}\phantomcaption\label{fig:KSE_ExampleControlledTraj-a}
	\end{subfigure}
	\begin{subfigure}[t]{0.0\textwidth}
		\includegraphics[width=\textwidth]{KSE_RL_traj}\phantomcaption\label{fig:KSE_ExampleControlledTraj-b}
	\end{subfigure}
	\begin{subfigure}[t]{0.0\textwidth}
		\includegraphics[width=\textwidth]{KSE_RL_traj}\phantomcaption\label{fig:KSE_ExampleControlledTraj-c}
	\end{subfigure}
	\begin{subfigure}[t]{0.0\textwidth}
		\includegraphics[width=\textwidth]{KSE_RL_traj}\phantomcaption\label{fig:KSE_ExampleControlledTraj-d}
	\end{subfigure}
	\begin{subfigure}[t]{0.0\textwidth}
		\includegraphics[width=\textwidth]{KSE_RL_traj}\phantomcaption\label{fig:KSE_ExampleControlledTraj-e}
	\end{subfigure}
	\caption[]{ROM-based  RL agent applied to the same initial condition in the a) true KSE and b) data-driven reduced-order model (decoded, $d_h=12$). The corresponding invariant quantities of dissipation and total input power for the c) true KSE and d) learned reduced-order model. The dashed black line represents the system average of the natural KSE dynamics. e) Controlled dissipation trajectories of the true KSE beginning from 15 randomly sampled test initial conditions of the KSE.}
	\label{fig:KSE_ExampleControlledTraj-full}
\end{figure}

To assess the performance of our \KZrevise{DManD-RL} policy, the learned control agent was applied to the NODE-ROM, with an example controlled trajectory shown in Fig. \ref{fig:KSE_ExampleControlledTraj-a}. We note that after a brief control transient, the control agent navigates the NODE-ROM to an equilibrium(steady) state and stabilizes it. The quantities targeted for minimization, $D$ and $P_f$, estimated from the {predicted} trajectory {$u(t)$}, are shown in Figure \ref{fig:KSE_ExampleControlledTraj-c}, revealing that this equilibrium  exhibits dissipation much lower than the natural unactuated dynamics. To assess how well this \KZrevise{DManD-RL} control policy {transfers} to the original KSE (i.e.~the true system), the same policy is applied to the true KSE with the same initial condition, as shown in Fig. \ref{fig:KSE_ExampleControlledTraj-b}. We note that the controlled trajectory in the KSE yields not only quantitatively similar transient behavior but also the same low-dissipation equilibrium state as was targeted in the NODE-ROM. The transient behaviors between the two are structurally very similar, although the NODE-ROM {displays slightly less strongly damped oscillations as it drives the trajectory to the steady state}. The values of $D$ and $P_f$ computed from the true system, shown in Fig. \ref{fig:KSE_ExampleControlledTraj-d}, are nearly identical to that of the NODE-ROM in Fig. \ref{fig:KSE_ExampleControlledTraj-c}. 

To demonstrate the robustness of the \KZrevise{DManD-RL} policy, shown in Fig.~\ref{fig:KSE_ExampleControlledTraj-e} are the dissipation trajectories of the true KSE beginning from 15 randomly sampled test initial conditions that the \KZrevise{DManD-RL} control agent has not seen before. We highlight that the control agent is able to consistently navigate the system to the same low-dissipation state within $\sim 150$ time units, with one initial condition requiring $\sim 200$ time units to converge. Finally we note that although the RL training horizons were only 100 time units long, the control agent is able to generalize to achieve and maintain control well beyond the the horizon it was trained in.

Here we emphasize that the \KZrevise{DManD-RL} policy drives the dynamics to an equilibrium state in both the NODE-ROM and the true KSE, indicating that not only does the NODE-ROM capture this state, but it captures the dynamics leading to it accurately enough such that the RL agent could discover it during training and exploit it in a manner that still translates to the original system. We further emphasize that both the NODE-ROM and agent were never explicitly informed of this low-dissipation state's existence. \KZrevise{Finally, we highlight that the discovered low-dissipation state is an unstable state that is stabilized by the control agent. If control is removed, the system returns to the natural chaotic dynamics}.

{These observations indicate that the RL policy trained on the model transfers very well to the true system.} We attribute this performance to the fact that both the NODE-ROM and RL agent operate in Markovian fashion, i.e. even if the model has {slight inaccuracies}, so long as the modeled dynamics are reasonably accurate {this does not matter} once the agent makes its new state observation. 

\begin{figure}[t]
	\begin{center}
		\includegraphics[width=0.5\textwidth]{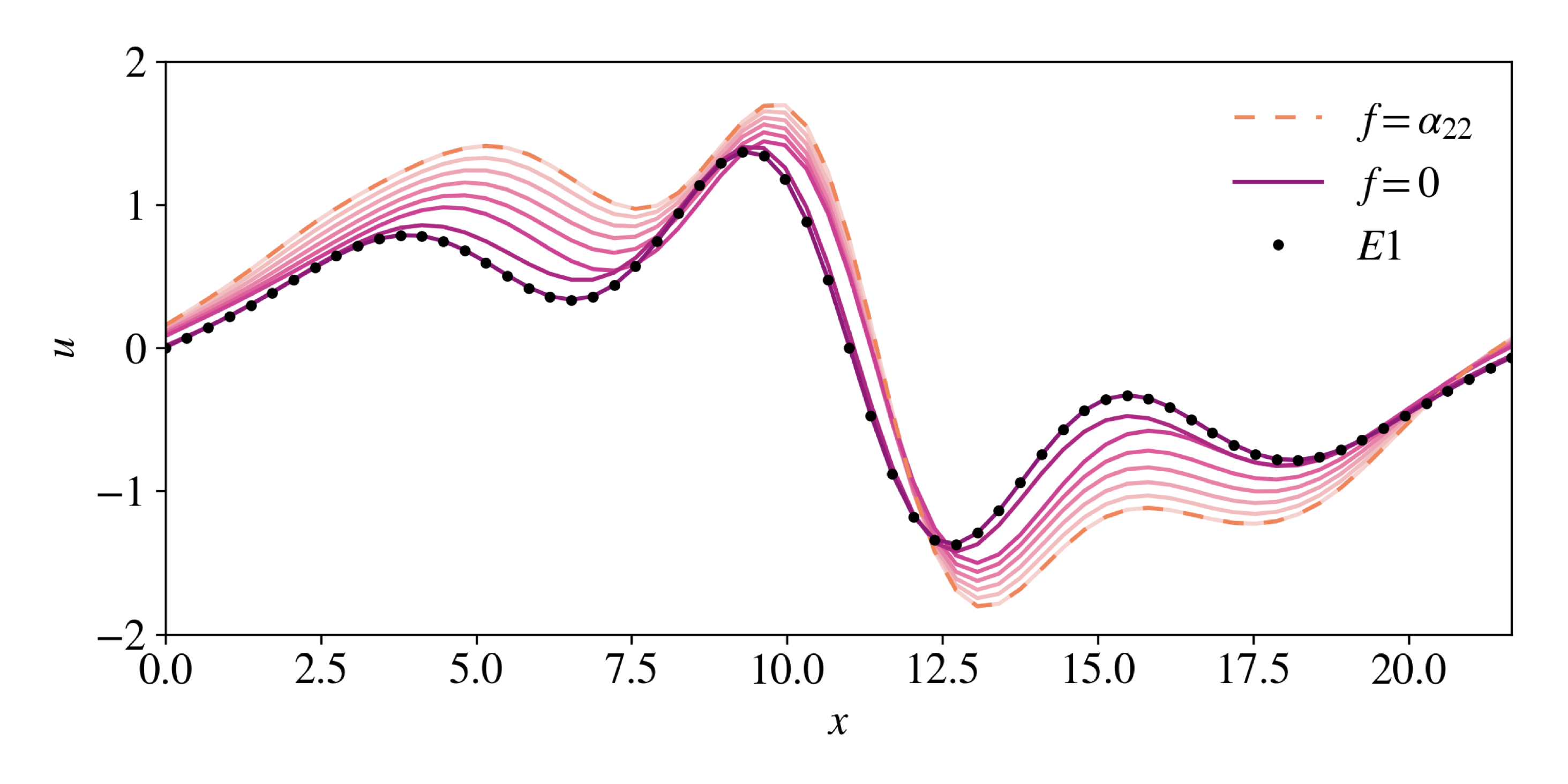}
		\caption[]{Forcing continuation from the forced equilibrium state (under forcing $f={\alpha_{22}}$) discovered by NODE-ROM based RL policy (orange, dashed) to the unactuated KSE system (purple). The known equilibrium E1 of the KSE system is also provided (dots).}
		\label{fig:KSE_Continuation}
	\end{center}
\end{figure}

{Returning to} the dynamical significance of the low-dissipation equilibrium state discovered and stabilized by the RL agent, a continuation in mean forcing magnitude was performed. To do so, we Newton-solved for equilibrium solutions to the KSE starting with the discovered equilibrium state while gradually decreasing the magnitude of the mean actuation profile to zero, as was done in \citep{Zeng2021}. Solutions identified by this continuation in forcing magnitude are shown in Fig. \ref{fig:KSE_Continuation}, which reveals that equilibrium state captured by the NODE-ROM and discovered by the RL agent is connected to a known existing solution of the KSE known as $E1$ \citep{Cvitanovic2010}; {we obtained a similar result with an RL agent trained on \KZrevise{interactions with the full system \citep{Zeng2021}}}. {A similar observation was made for RL control of 2D bluff body flow \citep{LiZhang2022}}. { We speculate that in systems with complex nonlinear dynamics, the discovery and stabilization of desirable underlying equilibrium solutions (or other recurrent saddle-point solutions such as unstable periodic orbits) of the system may be a fairly general feature of RL flow control approaches. The nonlinear and exploratory nature of RL algorithms facilitates the discovery of such solutions, and since \KZrevise{the dynamics are slow near these solutions}, little control action should be required to keep trajectories near them.}

We speculate that in systems with complex dynamics, the discovery and stabilization of desirable underlying solutions of the system may be {a fairly general feature of RL flow control approaches that aim to minimize dissipation while penalizing control action -- }; as was shown with the KSE in \citep{Zeng2021} and bluff-body flows in \citep{LiZhang2022}. This is a promising outlook for RL as the dynamics of even more complex chaotic systems, such as turbulent flows governed by the Navier-Stokes equations, are also known to be structured around such solutions \citep{FloryanReview2021}.


\KZrevise{To highlight the importance of the type of data-driven model, we compare to Dynamic Mode Decomposition (DMD), a common data-driven method that has been applied to fluid flows and dynamical systems. Recently, Qin et al. \citep{Qin2021} demonstrated active flow control of 2D cylinder flow with a reward signal targeting the minimization of DMD mode amplitudes. Here we specifically investigate the performance of using DMDc \citep{Proctor2016}, the extension of DMD to account for control input, as an alternate data-driven model of the KSE dynamics for RL policy learning.} The DMDc model takes the following form,
\begin{equation}
  s_{t+\tau} = \mathbf{A}s_t + \mathbf{B}a_t,
 \label{eq:DMDc}
\end{equation}
where $\mathbf{A}$ is the state transition matrix and $\mathbf{B}$ is the control input matrix. Using the same training data as our NODE-ROM, $[s_t, a_t, s_{t+\tau}]$, , we obtained a full state DMDc model by simultaneously fitting the two matrices with the $[s_t, a_t, s_{t+\tau}]$ data as described in \citep{Proctor2016}. {This model was then used as the environment for the RL policy training.} 

\begin{figure}[t]
 	\centering
	\begin{subfigure}[t]{0.9\textwidth}
		\includegraphics[width=\textwidth]{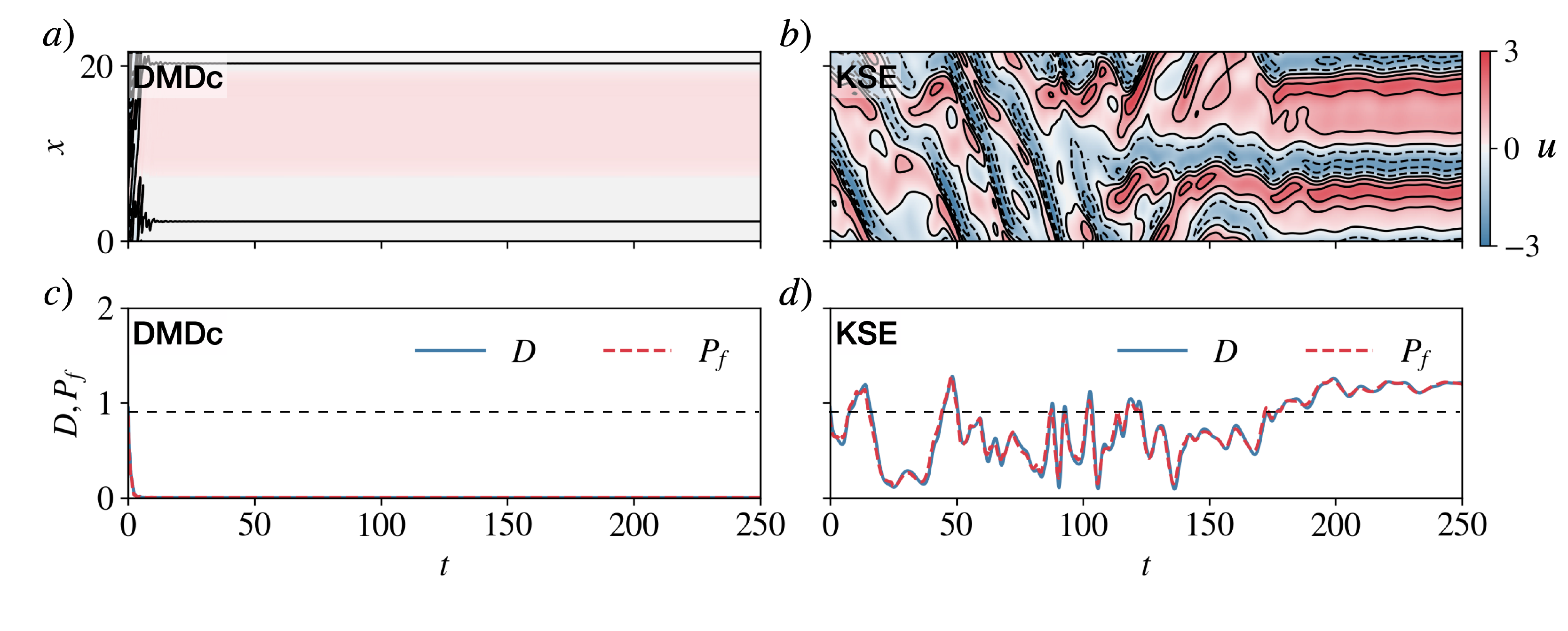}\phantomcaption\label{fig:KSE_ExampleDMDcTraj-a}
	\end{subfigure}
	\begin{subfigure}[t]{0.0\textwidth}
		\includegraphics[width=\textwidth]{KSE_DMDc_Traj}\phantomcaption\label{fig:KSE_ExampleDMDcTraj-b}
	\end{subfigure}
	\begin{subfigure}[t]{0.0\textwidth}
		\includegraphics[width=\textwidth]{KSE_DMDc_Traj}\phantomcaption\label{fig:KSE_ExampleDMDcTraj-c}
	\end{subfigure}
	\begin{subfigure}[t]{0.0\textwidth}
		\includegraphics[width=\textwidth]{KSE_DMDc_Traj}\phantomcaption\label{fig:KSE_ExampleDMDcTraj-d}
	\end{subfigure}
	\caption[]{DMDc control policy applied to the same initial condition in the (a) DMDc model and (b) True KSE. Corresponding dissipation and total input power for the (c) DMDc model and (d) KSE. The dashed black line represents the  average of the natural KSE dynamics. }
	\label{fig:KSE_ExampleDMDcTraj-full}
\end{figure}

\KZrevise{Shown in Fig. \ref{fig:KSE_ExampleDMDcTraj-a} is the DMDc-based RL control agent applied to the DMDc model it was trained in. We show in Fig. \ref{fig:KSE_ExampleDMDcTraj-c} that the agent is able to rapidly minimize the DMDc model values of $D, P_f$. We highlight that the DMDc-based strategy accomplishes this by exploiting the unphysical dynamics of the DMDc model. To elaborate, shown in Fig. \ref{fig:KSE_ExampleDMDcTraj-b} is the same strategy applied to the true KSE with the same initial condition as the DMDc trajectory in Fig. \ref{fig:KSE_ExampleDMDcTraj-a}. The DMDc-based RL strategy fails to minimize the control objective when applied to the true KSE. Furthermore, the strategy even results in  performance that is even worse than the time-averaged uncontrolled system, which can be seen in the above average $D, P_f$ values at long times in Fig. \ref{fig:KSE_ExampleDMDcTraj-d}, where the controller appears to get trapped in a high energy region of the KSE. This mismatch in performance is due to the inability of the linear model to describe the nonlinear dynamics of the KSE.}

\section{Conclusions} \label{Conclusions}
In this paper we introduce our NODE-ROM-Based RL method, DManD-RL, that maintains an end-to-end data-driven method of learning control policies from a limited data set. The governing equations and control term do not need to be known or specified--we only assume that the system's actuated dynamics can be represented by some governing system of ODEs in the  manifold coordinate. {\KZrevise{Furthermore, we do not impose any structural relationship between the control input and the dynamics of the systems as previous methods have done in the past. This is advantageous for many complex flow control systems as the relationship between the dynamics and the control inputs are not simple linear relationships \citep{Boskovic2004,Geranmehr2015}}. We exploit the notion that many nominally high-dimensional systems actually display low-dimensional dynamics and aim to capture and model these actuated dynamics from data using a combination of autoencoders and Neural ODEs. Then, with deep RL, we extract control policies from our data-driven ROMs. Importantly, our method does not require RL to directly interact with the target system nor does it require great modification of standard RL algorithms.


Our method identifies the lower-dimensional manifold the actuated dynamics live on with autoencoders  and models the dynamics with neural ODEs, which are capable of making  predictions for arbitrary time intervals \citep{Chen2018} and have been demonstrated to produce good predictions of spatiotemporal chaotic systems \citep{Linot2021}. Neural ODEs are also a natural formulation and deep architecture for the systems we are attempting to model and control. Our method also respects the Markovian nature underlying many RL frameworks as well as those of our target dynamical systems, unlike other common data-driven methods such as reservoir computing and recurrent neural networks. 
The use of high-fidelity low-dimensional models allows for faster interaction simulations (or even running multiple surrogate environments in parallel) compared to costly simulations or experiments.

By using a more informative and compact state space representation than a high-dimensional sensed state, the RL policy training load is lessened. In naive applications of RL, a portion of the agent's network capacity are exhausted in learning to transform the raw inputs to more useful internal representations \citep{Ha2018}. Similar to Ha and Schmidhuber \citep{Ha2018}, in our method we also explicitly separate the agent's learning of the control policy and the task of learning a useful state representation into two discrete tasks.

We apply our \KZrevise{DManD-RL} RL method to the KSE, a proxy system for turbulent flows that exhibits rich spatiotemporal chaotic dynamics. From a limited data set generated from trajectories of the KSE under random actuations, we are able to find a reduced-order mapping between the full state and the reduced low-order dynamics of the actuated KSE, as well as to successfully model the dynamics in this reduced space. We find that our NODE-ROM has good forecasting ability and captures ensemble characteristics of the KSE well for not only actuated predictions, but also unactuated predictions, for which it was not given training data. 
In this work it was found that NODE-ROM training data obtained from snapshots of the KSE experiencing random jet actuations was sufficient. 
With even more complex systems, it may be possible that this sampling approach is insufficient. If necessary, the NODE-ROM can be iteratively improved by applying the learned control strategy to the true system, collecting additional data, and fine-tuning the NODE-ROM with the new data. Using the updated NODE-ROM, the RL agent can then be updated. Together the control agent and model can be improved in a cyclic training fashion. \KZrevise{Finally, we highlight that our model could be alternatively trained and used in the ``Dyna"-style \citep{SuttonBarto2018} for improved online learning.}

Utilizing this NODE-ROM in place of the environment, the RL agent is able to discover a low-dissipation equilibrium state and learns to exploit it to minimize $D,P_f$. When the \KZrevise{DManD-RL} control policy is deployed to the true KSE, we observe that not only are the same equilibrium states targeted, but the performance is nearly indistinguishable from our learned NODE-ROM. This indicates that not only does the NODE-ROM capture the existence of the forced equilibrium, but it also captures the dynamics sufficiently well such that the agent could find it during training and exploit it in a manner that still translates in the original system. We emphasize that we accomplished this with a 12-dimensional NODE-ROM while the full state is 64- dimensional. A continuation in the magnitude of the forcing profile reveals that the RL discovered equilibrium state is connected to an existing equilibrium solution of natural KSE. The naturally occurring RL optimization about underlying solutions of the system has been observed in the KSE \citep{Zeng2021} as well as bluff-body flows \citep{LiZhang2022}, which is promising, as more complex dissipative systems, such as the turbulent dynamics of the Navier-Stokes equations, are also known to be organized about various types of invariant solutions \citep{FloryanReview2021}.

\end{document}